\documentclass{article}
\usepackage{hyperref}

\usepackage{microtype}
\usepackage{graphicx}
\usepackage{subfigure}
\usepackage{amsmath}
\usepackage{booktabs} 
\usepackage{amsmath}
\usepackage{cuted}
\usepackage{tabularx}
\usepackage{array}    
\usepackage{ragged2e}   
\usepackage{enumitem} 

\DeclareMathOperator*{\argmin}{arg\,min}

\usepackage[accepted]{icml2024}

\usepackage{amsmath}
\usepackage{amssymb}
\usepackage{mathtools}
\usepackage{amsthm}
\usepackage{tikz}
\usetikzlibrary{positioning,fit,calc,arrows.meta,shapes}
\usepackage[capitalize,noabbrev]{cleveref}

\theoremstyle{plain}

\theoremstyle{definition}

\theoremstyle{remark}

\usepackage[textsize=tiny]{todonotes}
\usepackage{hyperref}

\begin{document}

\twocolumn[
\icmltitle{Technical Report on Text Dataset Distillation}



\icmlsetsymbol{equal}{*}

\begin{icmlauthorlist}
\icmlauthor{Keith Ando Ogawa}{usp}
\icmlauthor{Bruno Lopes Yamamoto}{usp}
\icmlauthor{Lucas Lauton de Alcantara}{usp}
\icmlauthor{Victor Zacarias}{ime}
\icmlauthor{Edson Bollis}{icti}
\icmlauthor{Lucas Pellicer}{icti}
\icmlauthor{Rosimeire Pereira Costa}{icti}
\icmlauthor{Anna Helena Reali Costa}{usp}
\icmlauthor{Artur Jordao}{usp}
\end{icmlauthorlist}

\icmlaffiliation{usp}{Escola Politécnica, Universidade de São Paulo, São Paulo, Brazil}
\icmlaffiliation{ime}{Instituto de Matemática e Estatística, Universidade de São Paulo, São Paulo, Brazil}
\icmlaffiliation{icti}{Instituto de Ciência e Tecnologia Itaú (ICTi), São Paulo, Brazil}

\icmlcorrespondingauthor{}{}

\icmlkeywords{Machine Learning, ICML}

\vskip 0.3in
]



\printAffiliationsAndNotice{}  

\begin{abstract}
In the vision domain, dataset distillation arises as a technique to condense a large dataset into a smaller synthetic one that exhibits a similar result in the training process. While image data presents an extensive literature of distillation methods, text dataset distillation has fewer works in comparison. Text dataset distillation initially grew as an adaptation of efforts from the vision universe, as the particularities of the modality became clear obstacles, it rose into a separate branch of research. Several milestones mark the development of this area, such as the introduction of methods that use transformer models, the generation of discrete synthetic text, and the scaling to decoder-only models with over 1B parameters. Despite major advances in modern approaches, the field remains in a maturing phase, with room for improvement on benchmarking standardization, approaches to overcome the discrete nature of text, handling complex tasks, and providing explicit examples of real-world applications. In this report, we review past and recent advances in dataset distillation for text, highlighting different distillation strategies, key contributions, and general challenges.
\end{abstract}

\section{Introduction}
In the data-driven scenario of deep learning, dataset distillation emerges as a method to distill the knowledge of a large real dataset into a small synthetic one that is approximately equivalent in the training process~\cite{DD-Seminal:2020}. From a practical perspective, this technique improves dataset storage efficiency and reduces training time, contributing towards a more financially accessible and environmentally responsible way to develop machine learning models. Consequently, dataset distillation shows utility in applications that involve intensive training or constrained computational resources, such as neural architecture search~\cite{DD-NAS:2020} and on-device machine learning~\cite{MDS:2024}. From a theoretical perspective, the study of how to condense large amounts of data into a few samples has the potential to answer key questions about training models, such as “\emph{What defines good data}?”~\cite{softlabel:2024}.

Even though large language models (LLMs) play important roles in several recent breakthroughs~\cite{AiIndex:2025}, most of the literature on dataset distillation does not address them and concentrates on the vision domain~\cite{dd-survey-ijcai:2023}.

In this work, we aim to present not only the big picture of text dataset distillation research, but also the current state of art in the text domain. We organize our report as follows: Section~\ref{Sec:Taxonomy} introduces a taxonomy of methods; Section~\ref{Sec:StateOfArt} describes the current state of art, milestones and characteristics of text dataset distillation; Section~\ref{Sec:Challenges} addresses challenges and unexplored directions in text dataset distillation. Finally, Section~\ref{Sec:Conclusion} concludes this work with our impressions over the scarce literature.

\begin{figure}[t]
	\centering
	\includegraphics[width=\columnwidth]{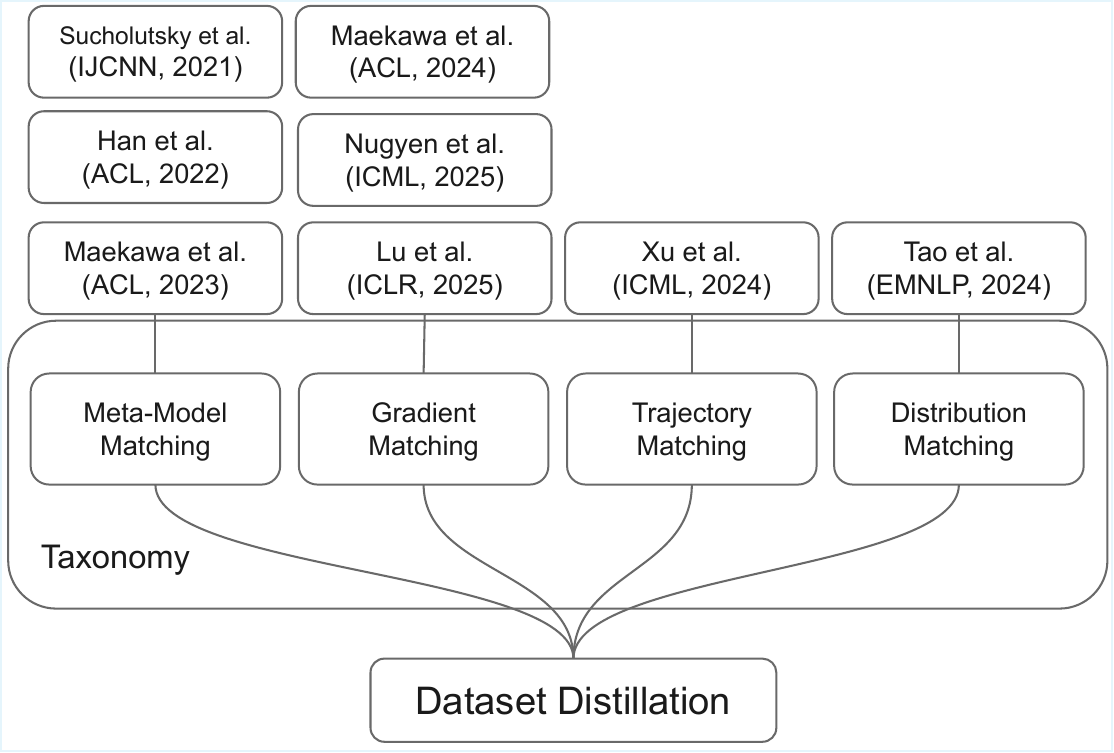}
    \vspace{-0.5cm}
	\caption{Taxonomy of Dataset Distillation Methods.}
	\label{Fig:Taxonomy}
    \vspace{-10pt}
\end{figure}

\section{Taxonomy}\label{Sec:Taxonomy}

In this report, we adopt the four-class taxonomy of dataset distillation methods proposed by Sachdeva et al.~\yrcite{DD-Survey:2023} (Figure~\ref{Fig:Taxonomy}): Meta-Model Matching, Gradient Matching, Trajectory Matching and Distribution Matching.

\subsection{Meta-Model Matching}

In their seminal work, Wang et al.~\yrcite{DD-Seminal:2020} coin the term "Dataset Distillation" to name the practice of distilling knowledge from a large training dataset into a small one. The authors solve dataset distillation through a bi-level optimization, in contrast with the training process, the proposed distillation optimization process tunes the synthetic data $\tilde{x}$. The objective of the optimization problem is to minimize the loss evaluated on real data of the model trained on synthetic data, as Equation~\ref{eq:dd} shows:
\begin{equation}
\argmin_{\tilde{x}} \ell({x, \theta_0 - \eta \nabla_{\theta_0} \ell({\tilde{x}, \theta_0})}).
\label{eq:dd}
\end{equation}
Specifically, given a loss function $\ell$, the solution to the problem is the synthetic data $\tilde{x}$ that minimizes $\ell$ over the real data $x$ with the adjusted weight $\theta_0 - \eta \nabla_{\theta_0} \ell({\tilde{x}, \theta_0})$.

We consider Meta-Model Matching methods to be any dataset distillation technique that follow this framework of modeling the problem as a consecutive iteration to make the synthetic data match the performance of the real data on a specific loss.

\subsection{Gradient Matching}
In contrast with Meta-Model Matching techniques, Gradient Matching methods distill data by forcing the gradient calculated on the synthetic data to be similar to the real one~\cite{GradMatch:2021}. Formally, given a distance function $D$, the objective is to minimize the difference between the gradient for the loss $\ell$ with respect to the model parameters $\theta$ over the synthetic data $\tilde{x}$ and real data $x$, as Equation~\ref{eq:grad} shows:
\begin{equation}
\argmin_{\tilde{x}} \space D(\nabla_{\theta}\ell({\tilde{x},\theta}),\nabla_{\theta}\ell({x, \theta})).
\label{eq:grad}
\end{equation}
An important characteristic of the Gradient Matching approach is that, due to its simpler problem formulation, it presents a lower computational cost compared to Meta-Model Matching.
\subsection{Trajectory Matching}

Trajectory Matching methods aim to generate distilled data that leads to similar long-range learning dynamics of a train with real data~\cite{Traj:2022}. To achieve this objective, these techniques compare segments of the parameter trajectory of training with synthetic data to the real data training trajectory.
\begin{equation}
    \argmin_{\tilde{x}} \space {E(\tilde{\theta}_N,\theta_M)}.
    \label{eq:traj}
\end{equation}
The previous Equation~\ref{eq:traj} shows how the approach is mathematically described at a distillation step. Specifically, given $\theta_0$ as the initial parameter configuration, $\tilde{\theta}_N$ as parameters updated using synthetic data for N steps, $\theta_M$ as parameters updated using real data for M steps and $E$ as a error function, the distillation process objective is to minimize the difference, or error, between the parameters trained with the real and synthetic data.

Apart from theoretical details, as these techniques incorporate information about the training trajectory, their computational cost is naturally higher than Gradient  Matching methods that observe one step at a time.

\subsection{Distribution Matching}

Distribution Matching techniques build upon the hypothesis that datasets with similar distributions, according to a particular metric, lead to similar training processes~\cite{DD-Survey:2023}. Equation~\ref{eq:dist} describes the core purpose of these techniques. Specifically, given a distribution mismatch measure $D$, the goal is to find synthetic data $\tilde{x}$ that minimize the distance to real data $x$.
\begin{equation}
    \argmin_{\tilde{x}} \space {D(\tilde{x},x)}.
    \label{eq:dist}
\end{equation}
In contrast to other classes of methods, distribution matching follows a single-level optimization by design, which makes it a more scalable approach~\cite{DD-Survey:2023}.

\section{State of The Art, Milestones, Models and Datasets}\label{Sec:StateOfArt}
\begin{figure*}[!htb]
\centering
\includegraphics[width=1.0\linewidth]{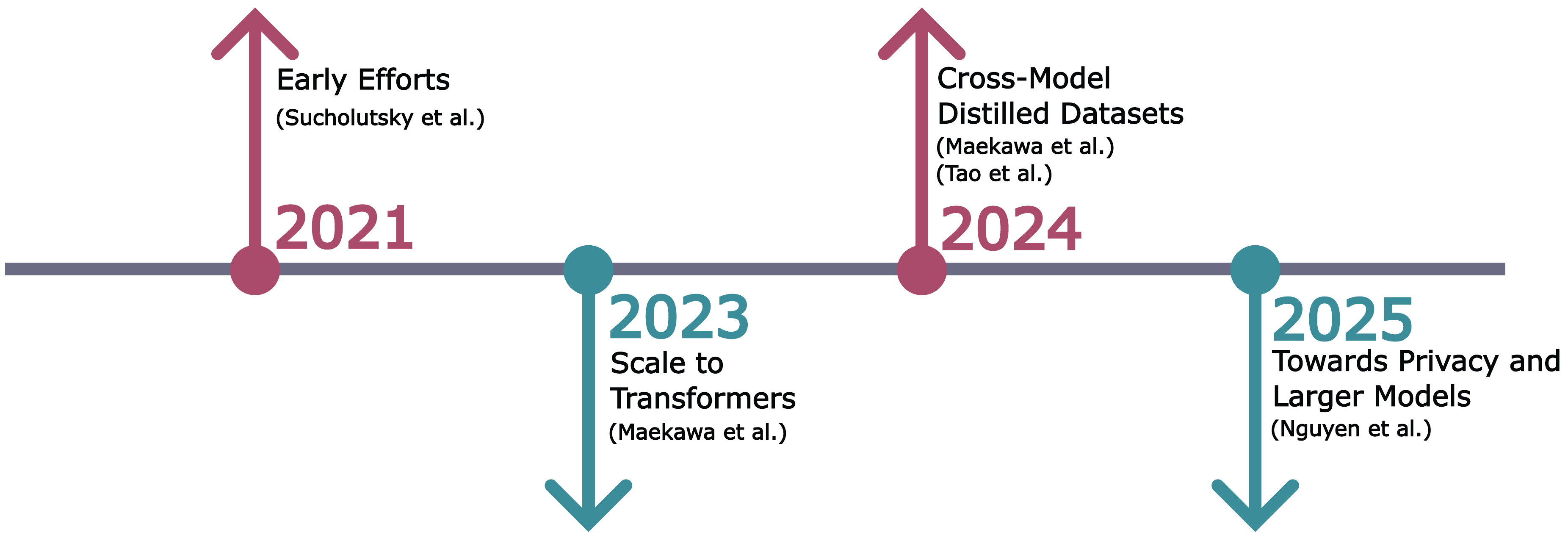}
\caption{Timeline of milestones in text dataset distillation. From 2021 to 2025, the timeline highlights important events that shape the research area. Evolving from a simple extension of a vision work, state-of-the-art dataset distillation methods now leverage transformer models of over a billion parameters to generate cross-architecture readable distilled datasets that guarantee privacy of the original data.}
\label{fig:time_line}
\end{figure*}

\subsection{Progression of Text Dataset Distillation Methods}

As an emerging research area, text dataset distillation evolved alongside image dataset distillation, incorporating and adapting its innovations. In the same work that introduces the use of soft labels, Sucholutsky et al.~\yrcite{DD-SoftLabels-Text:2021} introduce an early extension of these methods to the language domain. The authors successfully apply their use of label probability with the Meta-Model Matching approach to text embeddings, confirming that dataset distillation techniques are convertible to text data. Despite their theoretical and pioneering contributions, their work only addresses the distillation of classification datasets using CNN- and RNN-based architectures, which is inconsistent with modern use cases.

Given the importance of pre-trained large language models to natural language processing, Maekawa et al.~\yrcite{DD-AttentionLabels:2023} extend dataset distillation to encoder-only transformer models by tailoring special labels to this family of architectures: attention labels. 
The authors take inspiration from previous works that employ attention probabilities to supervise the training or knowledge distillation process. The idea is to create labels that incorporate this supervision as part of the synthetic data. More specifically, they initialize attention labels randomly and optimize them via minimization of the distance between their probabilities and attention probabilities over all layers and heads.

Together with attention labels, they use regular soft-labels in a Meta-Model Matching approach to generate distilled datasets. Their study with BERT~\cite{BERT:2019} on classification and natural language understanding tasks~\cite{GLUE:2018,AGNEWS:2015} show that the use of attention labels positively contributes to the performance of models trained with distilled data.

All previous work above generates synthetic datasets in the embedding space, thus making their techniques specific to models with the same embeddings. To achieve model-agnostic distilled datasets, the following recent studies aim to produce synthetic datasets in the discrete text space~\cite{DD-DILM:2024,DD-Embeddings:2024,DD-Syn:2025}.

Adopting approaches from the image domain, Maekawa et al.~\yrcite{DD-DILM:2024} tackle the problem through a different perspective; instead of optimizing discrete text, they shift the goal to train a model to generate the data via an easier continuous optimization. First, the authors pre-train a generator model to produce samples in the same distribution as the real data. Subsequently, they fine-tune this model to generate enhanced data using a gradient matching loss of the synthetic and real data through the learner model. Then, to create the synthetic dataset, they use the generator model to create up to a hundred times the desired amount of samples. Finally, by applying coreset selection methods, they obtain the final dataset of the desired size. Leveraging GPT-2 as the generator model and BERT as the learner model to calculate the gradients, they effectively synthesize a model-agnostic discrete dataset of classification tasks that outperforms coreset selection methods.

In a different direction, Tao et al.~\yrcite{DD-Embeddings:2024} approach the problem by using the embedding space to find and transform into text a small set of meaningful points that represent the real dataset. Rather than optimizing the data, the authors use embedding models to map the data into the embedding space, clustering algorithms to identify the synthetic dataset, and vec2text models to get the discrete form of the distilled dataset. Analysis over different embedding models, classification datasets, and models for evaluation shows that their technique effectively yields model-agnostic synthetic data.

Building upon the solid foundation laid by previous efforts, Nguyen et al.~\cite{DD-Syn:2025} propose a readable text generation technique that not only scales to distill data using large language models with a billion parameters but also guarantees the privacy of the original data. Featuring a gradient-matching approach, the authors employ the continuous nature of the embedding space to solve their optimization problem and project the synthetic embeddings into the discrete vocabulary space. In the optimization phase, their technique solves the gradient matching problem under the constraint that the solution needs to be in the set of all tokens via the Alternating Direction Method of Multipliers (ADMM). The authors claim to ensure differential privacy by injecting controlled noise into the real gradient during optimization, preventing inference of individual data points from the process output. In the projection phase, their method chooses tokens by selecting the one with the lowest Euclidean distance to the synthetic data. To generate readable text, only the top-k most probable tokens from the vocabulary are available on the projection phase. Finally, as the process is computationally intensive, they only match the gradient of the last layer of the model.

Their studies indicate the efficacy of the method, and their mathematical proofs guarantee the convergence of fine-tuning with the synthetic dataset. These results provide a stronger theoretical foundation for future research and demonstrate how to scale data distillation with privacy beyond encoder-only models, with the generated text also proving highly transferable to different LLM architectures.

\subsection{Fairness and Detoxification with Dataset Distillation}

Apart from performance evaluation, Han et al.~\yrcite{DD-fair:2022} focus on the fairness of distilled datasets. As the original dataset may have biases, it is intuitive to think that the distilled dataset could inherit them, and their experiments confirm it. To address this issue, the authors include bias mitigation mechanisms in their distillation process, successfully improving the fairness of the condensed data.

Specifically, they mitigate the inheritance of biases by both pre-processing the original dataset to balance the number of samples per class and modifying the optimization objective to include an adversarial training component. In particular, the latter part encourages the representations to be uninformative with respect to a specific target label. Their experiments show that the use of these mechanisms in the distillation process effectively boosts fairness with respect to a specific chosen attribute. As this work takes place in the early stage of text dataset distillation, its studies cover the use of a multi-layer perceptron on binary sentiment analysis and occupation classification tasks over datasets with bias~\cite{dataset-MOJI:2016,dataset-occupation-fair:2019}.

Addressing the challenge of toxicity in large language models, Lu et al.~\yrcite{DD-UNIDETOX:2025} introduce UniDetox, a framework centered on a novel and scalable dataset distillation method. Their approach leverages contrastive decoding to distill detoxifying knowledge into a concise, synthetic text dataset. They achieve detoxification by generating text that maximizes the difference in log probabilities between a base model and its counterpart fine-tuned on toxic data.

The circumvention of the costly second-order derivative calculations and gradient-based optimizations common in other distillation methods leads to a computationally efficient method that scales to large models. Furthermore, by distilling directly into discrete, human-readable text, UniDetox produces a universally applicable and tokenizer-agnostic dataset. This model-agnostic nature allows a single distilled dataset generated from a smaller model like GPT-2 to effectively detoxify a variety of larger models with different architectures, including OPT, Falcon, and LLaMA-2. Importantly, the detoxification through a simple fine-tuning process preserves perplexity and task predictive performance. The authors theoretically frame their method as an implicit form of gradient matching. In this approach, contrastive decoding samples texts whose gradients alignwith the opposite direction of the toxicity vector, thereby ppromotinglearning" of toxic behaviors.

\subsection{Extending to multimodal data}

Extending beyond the language domain, Xu et al.~\yrcite{DD-MULTIMODAL-SIMILARITY:2024} tackle multimodal dataset distillation, exploiting particular characteristics of image-text contrastive learning data. 

Directly applying dataset distillation to this type of data generates a synthetic dataset where images and text have a single right correspondence between them. In contrast to this standard approach, the authors propose to enhance the distilled data with learnable similarity information between each image-text pair. Therefore, instead of having a single right connection, every possible pair has a similarity value that measures how much the text matches the image.

They implement the generated similarity matrix as a sum of a diagonal component and a low-rank component, leading to a computationally feasible approach that successfully increases the performance of the base trajectory matching strategy on image-text data.

\subsection{Models and Datasets}

\begin{table*}[t]
\caption{Overview of taxonomies, works, source models (used for distillation), target models (used for evaluation), and datasets in dataset distillation.}
\smallskip\noindent
\resizebox{\textwidth}{!}{%
\begin{tabular}{p{4cm} p{5cm} p{5cm} p{5cm} p{5cm}}
\hline
Taxonomy              & Works                            & Source model & Target model & Dataset\\ \hline
Meta-Model Matching   & Sucholutsky et al. (IJCNN, 2021) & TextConvNet, Bi-RNN, Bi-LSTM & TextConvNet, Bi-RNN, Bi-LSTM  & IMDB dataset~\cite{dataset-imdb:2011}, SST5 dataset~\cite{dataset-sst5:2013}, TREC dataset~\cite{dataset-trec:2000} \\
Meta-Model Matching   & Han et al. (ACL, 2022)           & MLP  & MLP & MOJI dataset~\cite{dataset-MOJI:2016}, 28-way occupation classification dataset~\cite{dataset-occupation-fair:2019}\\
Meta-Model Matching   & Maekawa et al. (ACL, 2023)       & BERT & BERT & AGNews~\cite{AGNEWS:2015}, SST2, MRPC, QNLI~\cite{GLUE:2018}\\ \hline
Gradient Matching     & Maekawa et al. (ACL, 2024)       & GPT2, BERT   & BERT, RoBERTa, XLNet & SST2, QQP, MNLI-m~\cite{GLUE:2018} \\
Gradient Matching     & Nguyen et al. (ICML, 2025)       & Phi-1.5   &  Llama-3.2-1B, OPT-1.3B, Phi-1.5 & SST2~\cite{GLUE:2018}, Tweet Emotions dataset~\cite{dataset-tweet:2018}, Rotten Tomatoes dataset~\cite{dataset-rotten-tomatoes:2005}\\
Gradient Matching     & Lu et al. (ICLR, 2025)           & GPT2-XL &  GPT2-XL, OPT-6.7B, Falcon-7B, LLaMA2-7B & Dinamically Generated Hate Speech~\cite{dataset-hate-speech:2021}, ToxiGen~\cite{dataset-toxigen:2022} \\ \hline
Trajectory Matching   & Xu et al. (ICML, 2024)           & NFNet + BERT  & RestNet50 + BERT, RegNet + BERT, NFNet + BERT, ViT + BERT, NFRegNet + BERT, NFResNet + BERT,  NFNet + DistilBERT, NFNet + CLIP-Text  & Flickr30k dataset~\cite{dataset-flickr30k:2015}, COCO~\cite{dataset-coco:2014} \\ \hline
Distribution Matching & Tao et al. (EMNLP, 2024)         & GloVe, e5-base-v2, text-embedding-ada-2-002, text-embedding-large-3, T5-Base   &    Logistic Regression, Naive Bayes, Support Vector Machine, TextRNN, TextCNN, BERT, T5-Base  & IMDB dataset, AGNews~\cite{AGNEWS:2015}\\ \hline
\end{tabular}
}
\label{tab:models_and_datasets}
\end{table*}

\textbf{Datasets.}
The majority of the works we review assess their distillation methods with text classification datasets. We observe that the area not only limits the applications of techniques around this task, but also is devoid of standardization in terms of using specific datasets to ensure direct comparisons. Even though subsets of the GLUE benchmark~\cite{GLUE:2018} and AGNews~\cite{AGNEWS:2015} dataset appear frequently, most of the comparisons are only made with the seminal works. Table~\ref{tab:models_and_datasets} shows the datasets and their frequency of use in the work we cover, providing a reference for future research on dataset distillation.

\textbf{Models.} Early efforts employ tiny models, as RNNs, CNNs, and MLPs,  to distill text datasets~\cite{DD-SoftLabels-Text:2021,DD-fair:2022}. As the field gains maturity, Maekawa et al.~\yrcite{DD-AttentionLabels:2023} introduce text dataset distillation with the transformer architecture. Although recent works leverage modern decoder-only models~\cite{DD-UNIDETOX:2025,DD-Syn:2025}, there are currently no techniques that use architectures with more than 1.5B parameters to distill data. Table~\ref{tab:models_and_datasets} shows the model used in each work, offering an overview to better understand the area and support model selection in future work.

\section{Challenges}\label{Sec:Challenges}

Dataset distillation for language models represents a promising approach for compressing training data and reducing training costs. However, according to our review, several challenges remain, and the field is still incipient in comparison with computer vision. Current distillation approaches consider mostly small models (e.g. BERT and GPT-2) on simple datasets that represent problems very distant from current state-of-the-art use cases of LLMs. Modern datasets for fine-tuning language models (such as Omni-MATH~\cite{dataset-omnimath:2025}) contain more than simple classification examples, while current dataset distillation approaches mostly focus on this scope. Advancing into fine-tuning models with complex datasets could reveal many unexplored deficiencies of dataset distillation techniques, since the concept of an ideal distilled dataset could look very different when going beyond classification tasks. 

Although the literature reports practical reduction in dataset size, sometimes with minimal performance degradation, examples of real-world applications that account for total costs of running these methods are still missing. Another challenge regards data privacy—one of the main objectives of dataset distillation. More concretely, literature still lacks experiments validating the approaches on differential privacy scenarios.

Finally, even though being able to distill datasets with modern larger models is clearly positive to the research area, dataset distillation works do not highlight their practical benefits and improvements over datasets distilled with other models. Beyond this major gap, studies lack benchmark standardization, which harms the quantification of any novelty.

\section{Conclusions}\label{Sec:Conclusion}
In this work, we dive into the emerging area of text dataset distillation, covering a taxonomy to support our perception of the area. We also present a walkthrough of its milestones and current scope, and our perspective on the challenges within the research domain.

Our report shows that, although works in the language domain represent a minor portion of the dataset distillation research, recent punctual contributions made a substantial impact, pushing the field towards a direction where not only practical applications are viable, but also theoretical works have a solid foundation to develop on. Despite these important advances, we highlight that the area still faces several challenges, including overlooking practical applications, the absence of benchmark standards, and restrictions to classification tasks. 

Therefore, the area is still maturing but has promising avenues to explore with a better foundation due to the successful previous efforts. We believe that the advent of dataset distillation for large language models~\cite{DD-AttentionLabels:2023} and cross-architecture dataset distillation~\cite{DD-Syn:2025,DD-Embeddings:2024,DD-DILM:2024} lays the groundwork for theoretical and practical works that align with current AI challenges~\cite{AiIndex:2025}. In particular, taking a step into complex tasks, scaling to larger models, and making LLMs development more sustainable.

\section{Acknowledgments}

The authors would like to thank Instituto de Ciência e Tecnologia Itaú (ICTi) and Programa de Bolsas Itaú (PBI). This study was financed in part by the Coordenação de Aperfeiçoamento de Pessoal de Nível Superior – Brasil (CAPES) – Finance Code 001. Artur Jordao Lima Correia would like to thank Edital Programa de Apoio a Novos Docentes 2023. Processo USP nº: 22.1.09345.01.2. Anna H. Reali Costa would like to thank grant \#312360/2023-1 CNPq.

\bibliography{refs}
\bibliographystyle{icml2024}

\end{document}